\documentclass[11pt,a4paper]{article}
\usepackage[hyperref]{emnlp-ijcnlp-2019}
\usepackage{times}
\usepackage{latexsym}
\usepackage{graphicx}
\usepackage{amsmath}
\usepackage{breqn}

\usepackage{url}
\usepackage{balance}

\usepackage[margin=0.5cm]{subcaption}

\usepackage{amsmath,amsfonts,bm}

\def\eqref#1{equation~\ref{#1}}

\def\1{\bm{1}}

\def\vtheta{{\bm{\theta}}}

\def\vi{{\bm{i}}}

\def\vo{{\bm{o}}}

\def\vy{{\bm{y}}}

\DeclareMathAlphabet{\mathsfit}{\encodingdefault}{\sfdefault}{m}{sl}
\SetMathAlphabet{\mathsfit}{bold}{\encodingdefault}{\sfdefault}{bx}{n}

\usepackage{enumitem}
\setlist{noitemsep,topsep=2pt,parsep=2pt,partopsep=2pt}

\aclfinalcopy 

\usepackage{xcolor}

\title{EGG: a toolkit for research on Emergence of lanGuage in Games}
\author{Eugene Kharitonov \\
  Facebook AI \\
  {\tt kharitonov@fb.com} \\\And
  Rahma Chaabouni \\
  Facebook AI Research / LSCP \\
  {\tt rchaabouni@fb.com} \\\And
  Diane Bouchacourt \\
  Facebook AI \\
  {\tt dianeb@fb.com} \\\AND 
  Marco Baroni \\
  Facebook AI / ICREA \\
  {\tt mbaroni@fb.com}
}
\date{}

\begin{document}
\maketitle
\begin{abstract}
There is renewed interest in simulating language emergence among deep neural agents that communicate to jointly solve a task, spurred by the practical aim to develop language-enabled interactive AIs, as well as by theoretical questions about the evolution of human language. However, optimizing deep architectures connected by a discrete communication channel (such as that in which language emerges) is technically challenging. We introduce EGG, a toolkit that greatly simplifies the implementation of emergent-language communication games. EGG's modular design provides a set of building blocks that the user can combine to create new games, easily navigating the optimization and architecture space. We hope that the tool will lower the technical barrier, and encourage researchers from various backgrounds to do original work in this exciting area.
\end{abstract}
\section{Introduction}
\makeatletter{\renewcommand*{\@makefnmark}{}
\footnotetext{A screencast demonstration of EGG is available at \url{https://vimeo.com/345470060}}\makeatother}
Studying the languages that emerge when neural agents interact with each other recently became a vibrant area of research \cite{Havrylov2017, Lazaridou2016, Lazaridou2018, Kottur2017,Bouchacourt2018,Lowe2019}. Interest in this scenario is fueled by the hypothesis that the ability to interact through a human-like language is a prerequisite for genuine AI \cite{Mikolov2016,chevalier2019}. Furthermore, such simulations might lead to a better understanding of both standard NLP models~\cite{Chaabouni2019word} and the evolution of human language itself~\cite{Kirby2002}.


For all its promise, research in this domain is  technically very challenging, due to the discrete nature of communication. The latter prevents the use of conventional optimization methods,  requiring either Reinforcement Learning algorithms (e.g., REINFORCE; \citealt{Williams1992}) or the Gumbel-Softmax relaxation \cite{Maddison2016,Jang2016}. The technical challenge might be particularly daunting for researchers whose expertise is not in machine learning, but in fields such as linguistics and cognitive science, that could contribute to this interdisciplinary research area.

To lower the starting barrier and encourage high-level research in this domain, we introduce the EGG (\textbf{E}mergence of lan\textbf{G}uage in \textbf{G}ames) toolkit. EGG aims at\begin{enumerate}
    \item Providing reliable building bricks for quick prototyping;
    \item Serving as a library of pre-implemented games;
    \item Providing tools for analyzing the emergent languages.
\end{enumerate}

EGG is implemented in PyTorch~\citep{Paszke2017} and it is licensed under the MIT license. EGG can be installed from \url{https://github.com/facebookresearch/EGG}. 

Notable features of EGG include: (a) Primitives for implementing single-symbol or variable-length communication (with vanilla RNNs \cite{Elman:1990}, GRUs~\cite{Cho2014},  LSTMs~\cite{Hochreiter1997});\footnote{EGG also provides an experimental support of Transformers~\cite{Vaswani2017}.} (b) Training with optimization of the communication channel through REINFORCE or Gumbel-Softmax relaxation via a common interface; (c) Simplified configuration of the general components, such as check-pointing, optimization, Tensorboard support,\footnote{\url{https://www.tensorflow.org/tensorboard}} etc.; (d) A simple CUDA-aware command-line tool for hyperparameter grid-search.

\section{EGG's architecture}
In the first iteration of EGG, we concentrate on a simple class of games, involving a single, unidirectional (Sender $\to$ Receiver) message. In turn, messages can be either single-symbol or multi-symbol variable-length sequences. 
Our motivation for starting with this setup is two-fold. First, it corresponds to classic signaling games \cite{Lewis:1969}, it already covers a large portion of the literature (e.g., 5 out of 6 relevant studies mentioned in Introduction) and it allows exploring many interesting research questions. Second, it constitutes a natural first step for further development; in particular, the majority of components should remain useful in multi-directional, multi-step setups.



\subsection{Design principles}
As different  training methods and architectures are used in the literature, our primary goal is to provide EGG users with the ability to easily navigate the space of common design choices. 

Building up on this idea, EGG makes switching between Gumbel-Softmax relaxation-based and REINFORCE-based training effortless, through the simple choice of a different wrapper. Similarly, one can switch between one-symbol communication and variable-length messages with little changes in the code.\footnote{This also proved to be a convenient debugging mechanism, as single-symbol communication is typically simpler to train.}


We aim to maintain EGG minimalist and ``hackable'' by encapsulating
the user-implemented agent architectures, the Reinforce/GS agent wrappers and the game logic into PyTorch modules. The user can easily replace any part.

Finally, since virtually any machine-learning experiment has common pieces, such as setting the random seeds, configuring the optimizer, model check-pointing, etc., EGG pre-implements many of them, reducing the necessary amount of boilerplate code to the minimum. 

\subsection{EGG design}
EGG, in its first iteration, operates over the following entities. Firstly, there are two distinct agent roles: \textbf{Sender} and \textbf{Receiver}. Sender and Receiver are connected via a one-directional communication channel from the former to the latter, that has to produce the game-specific output.

\begin{figure}
\centering
\includegraphics[width=0.5\textwidth]{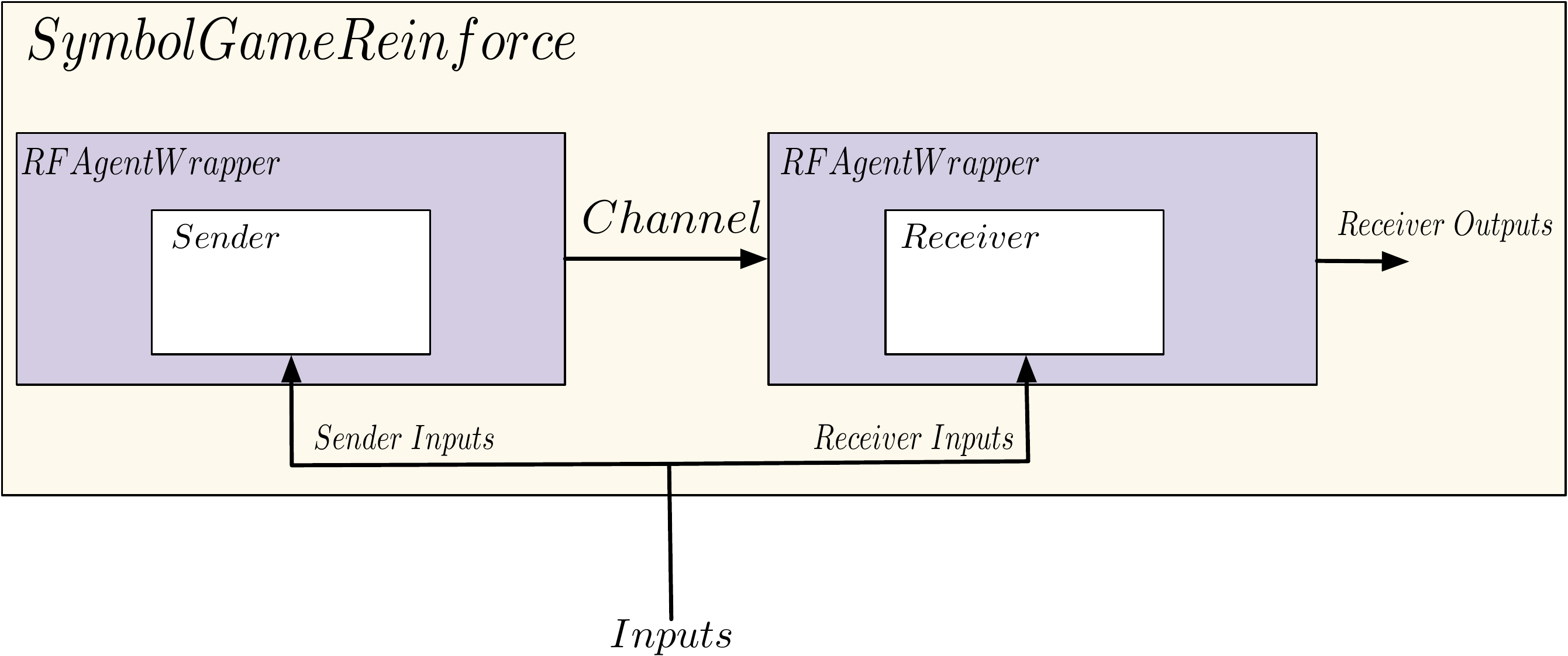}
\caption{Example of EGG's game flow when using REINFORCE. White boxes (Sender and Receiver) represent the user-implemented agent architectures. The colored boxes are EGG-provided wrappers that implement a REINFORCE-based scenario. For example, \textit{SymbolGameReinforce} is an instance of the Game block. It sets up single-symbol Sender/Receiver game optimized with REINFORCE.
To use Gumbel-Softmax relaxation-based training instead, the user only has to change the EGG-provided wrappers.}
\label{fig:example}
\end{figure}

The next crucial entity is \textbf{Game}. It encapsulates the agents and orchestrates the game scenario: feeding the data to the agents, transmitting the messages, and getting the output of Receiver. Figure \ref{fig:example} illustrates EGG's game flow in a specific example. Game applies a user-provided \textbf{loss} function, which might depend on the outputs of Receiver, the message transmitted, and the data. The value of the loss is minimized by a fourth entity, \textbf{Trainer}. Trainer also controls model checkpointing, early stopping, etc.

The Trainer and Game modules are pre-implemented in EGG. In a typical scenario, the communication method (single or multiple symbol messages) will be implemented by EGG-provided wrappers. As a result, what is left for the user to implement consists of: (a) the data stream, (b) core (non-communication-related) parts of the agents, (c) the loss.  The data interface that is expected by Trainer is an instance of the standard PyTorch data loader {\tt utils.data.DataLoader}.

To implement Sender, the user must define a module that consumes the data and outputs a tensor.
On Receiver's side, the user has to implement a module that takes an input consisting of a message embedding and possibly further data, and generates Receiver's output. 

Section \ref{sec:implementing-a-game} below provides examples of how to implement agents, choose communication and optimization type, and train a game.

\begin{figure}
\centering
\includegraphics[width=0.5\textwidth]{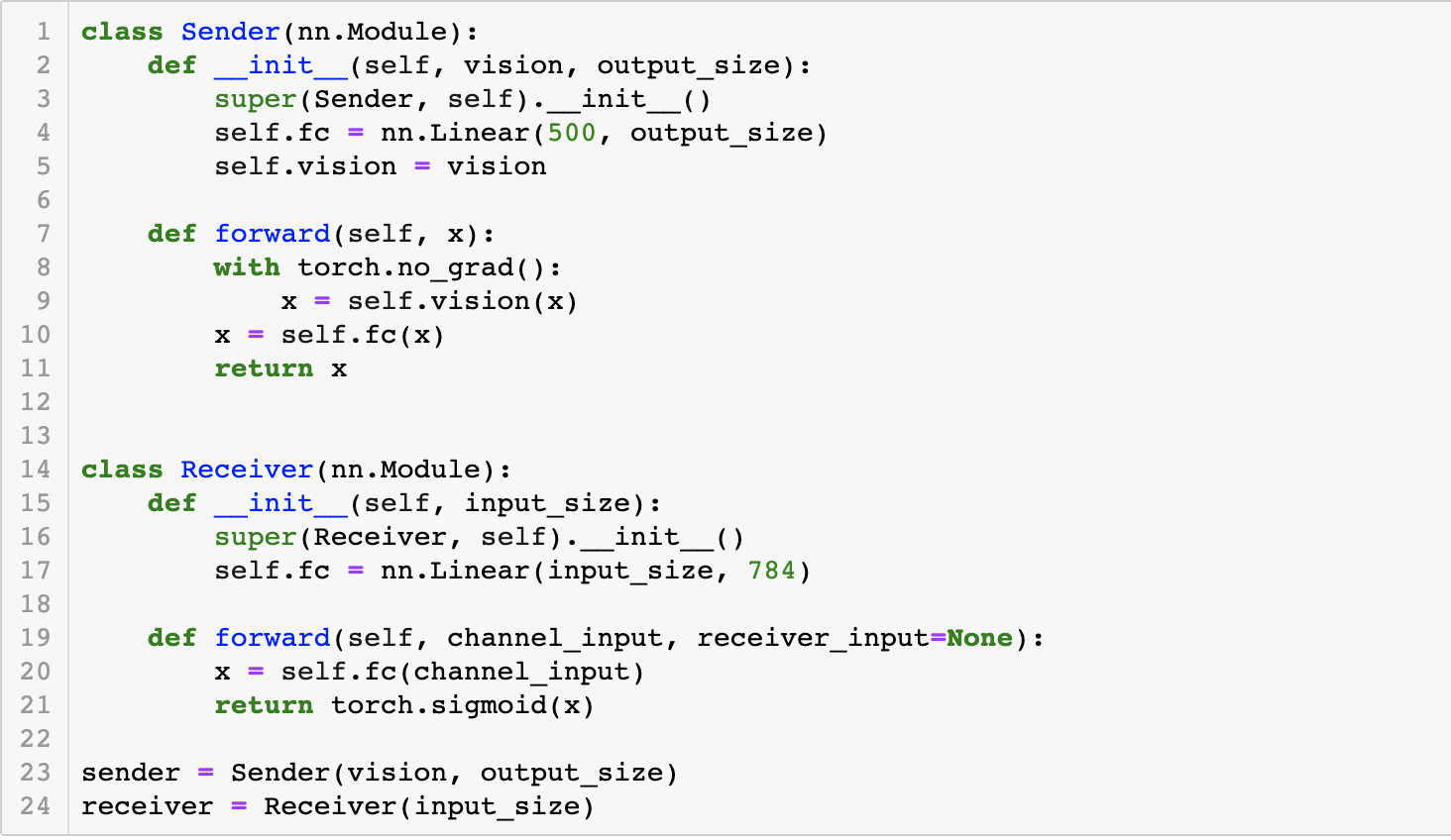}
\caption{MNIST game: Defining and instantiating the user-defined parts of the agents' architecture.}
\label{fig:agents}
\end{figure}

\begin{figure*}
  \centering
  \begin{subfigure}{0.5\textwidth}
  \centering
  \includegraphics[width=1.0\textwidth]{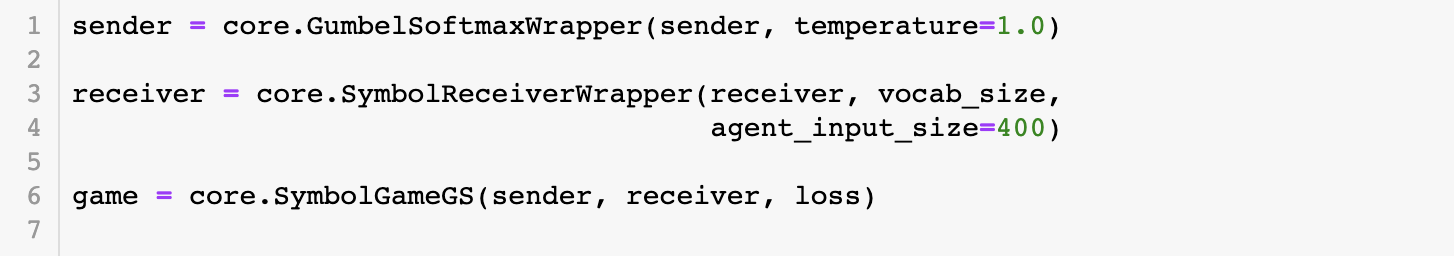}
  \caption{Single-symbol communication, Gumbel-Softmax relaxation.}
  \end{subfigure}%
    \begin{subfigure}{0.5\textwidth}
  \centering
  \includegraphics[width=1.0\textwidth]{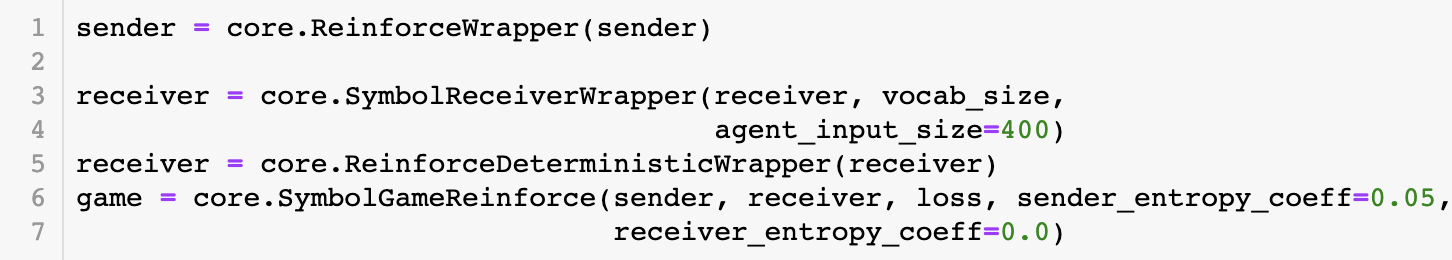}
    \caption{Single-symbol communication, REINFORCE.\newline}
  \end{subfigure}
\begin{subfigure}{0.5\textwidth}
  \centering
  \includegraphics[width=1.0\textwidth]{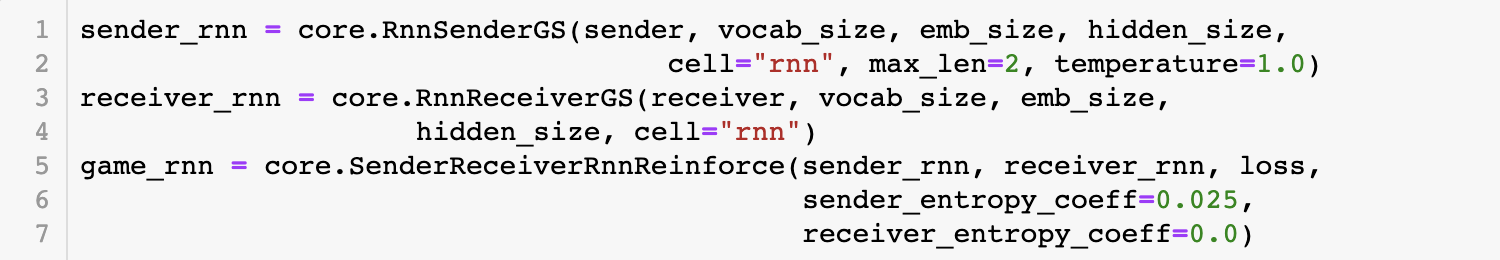}
      \caption{Variable-length communication, Gumbel-Softmax relaxation.}
    \end{subfigure}%
    \begin{subfigure}{0.5\textwidth}
  \centering
  \includegraphics[width=1.0\textwidth]{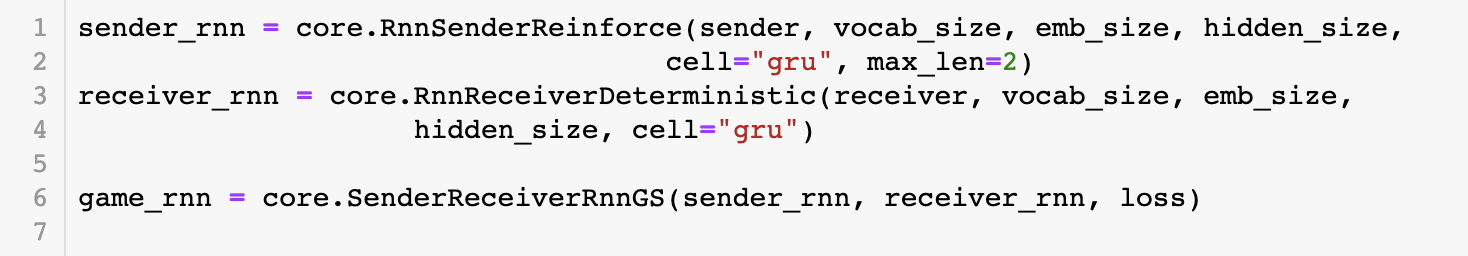}
  \caption{Variable-length communication, REINFORCE.\newline}
    \end{subfigure}%
\caption{MNIST game: The user can choose different communication wrappers to switch between training regimes (Gumbel-Softmax or REINFORCE) and communication type (single-symbol or variable-length messages).}
\label{fig:agent_wrappers}
\end{figure*}

\begin{figure}
\centering
\includegraphics[width=0.5\textwidth]{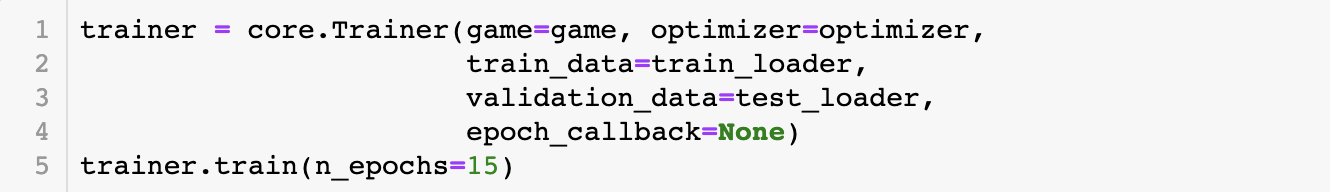}
\caption{MNIST game: Once the agents and the game are instantiated, the user must pass them to a Trainer, which implements the training/validation loop, check-pointing, etc.}
\label{fig:trainer}
\end{figure}



\section{Optimizing the communication channel in EGG} 
EGG supports two widely adopted strategies for learning with a discrete channel, Gumbel-Softmax relaxation (used, e.g., by \citet{Havrylov2017}) and REINFORCE (used, e.g., by~\citet{Lazaridou2016}). Below, we briefly review both of them.

\textbf{Gumbel-Softmax relaxation} is based on the Gumbel-Softmax (GS) (aka Concrete) distribution~\cite{Maddison2016,Jang2016}, that allows to approximate one-hot samples from a Categorical distribution. At the same time, GS admits reparametrization, hence allows backpropagation-based training. Suppose that Sender produces a distribution over the vocabulary, with $i$th symbol having probability $p_i = S(\vi_s)$. To obtain a sample from a corresponding Gumbel-Softmax distribution, we take i.i.d.\ samples $g_i$ from the {\tt Gumbel(0, 1)} distribution and obtain the vector $\vy$ with components $y_i$:
\begin{equation}
    y_i = \frac{exp((\log p_i + g_i)/\tau)}{\sum_j exp((\log p_j + g_j)/\tau)}
\end{equation}
where $\tau$ is the temperature hyperparameter, which controls the degree of relaxation. We treat $\vy$ as a relaxed symbol representation. In the case of single-symbol communication, the embedding of $\vy$ is passed to Receiver. In case of variable-length messages, the embedding is also fed into a RNN cell to generate the next symbol in the message. 

As a result, if Receiver and the game loss are differentiable w.r.t.\ their inputs, we can get gradients of all game parameters, including those of Sender, via conventional backpropagation.

\textbf{REINFORCE} \cite{Williams1991} is a standard Reinforcement Learning algorithm. Assume that both agents are stochastic: Sender samples a message $m$, and Receiver samples its output $\vo$.
Let us fix a pair of inputs, $\vi_s$, $\vi_r$, and the ground-truth output $l$. Then, using the log-gradient ``trick'', the gradient of the expectation of the loss $L$ w.r.t.\ the vector of agents' parameters $\vtheta = \vtheta_s \bigsqcup \vtheta_r$ is:
\begin{equation}
\mathbb{E}_{m, \vo}  \left[ L(\vo, l) \nabla_{\theta}  \log \mathbb{P}(m, \vo | \vtheta) \right]
\label{eq:reinforce}
\end{equation}
where $\mathbb{P}(m, \vo | \vtheta)$ specifies the joint probability distribution over the agents' outputs.

The gradient estimate is found by sampling messages and outputs. A standard trick to reduce variance of the estimator in Eq.~\ref{eq:reinforce} is to subtract an action-independent baseline $b$ from the optimized loss~\cite{Williams1992}. EGG uses the running mean baseline.

Importantly, the estimator in Eq.~\ref{eq:reinforce} allows us to optimize agents even if the loss is not differentiable (e.g., 0/1 loss). However, if the loss is differentiable and Receiver is differentiable and deterministic, this can be leveraged by a ``hybrid'' approach: the gradient of Receiver's parameters can be found by backpropagation, while Sender is optimized with REINFORCE. This approach, a special case of gradient estimation using stochastic computation graphs as proposed by~\citet{Schulman2015}, is also supported in EGG.

\section{Implementing a game}
\label{sec:implementing-a-game}
In this Section we walk through the main steps to build a communication game in EGG. We illustrate them through a MNIST~\cite{Lecun1998} communication-based autoencoding task: Sender observes an image and sends a message to Receiver. In turn, Receiver tries to reconstruct the image. We only cover here the core aspects of the implementation, ignoring standard pre- and post-processing steps, such as data loading. The full implementation can be found in an online tutorial.\footnote{ \url{https://colab.research.google.com/github/facebookresearch/EGG/blob/master/tutorials/EGG\%20walkthrough\%20with\%20a\%20MNIST\%20autoencoder.ipynb}} 

We start by implementing the agents' architectures, as shown in Figure~\ref{fig:agents}. Sender is passed an input image to be processed by its pre-trained {\tt vision} module, and returns its output after a linear transformation. The way Sender's output will be interpreted depends on the type of communication to be used (discussed below). Receiver gets an input from Sender and returns an image-sized output with pixels valued in $[0; 1]$. Again, depending on the type of channel employed, the Receiver input will have a different semantics.

In the case of one-symbol communication, Sender's output is passed through a {\tt softmax} layer and its output is interpreted as the probabilities of sending a particular symbol. Hence, the output dimensionality defines the size of the vocabulary. In the case of variable-length messages, Sender's output specifies the initial hidden state of an RNN cell. This cell is then ``unrolled'' to generate a message, until the end-of-sequence symbol ({\tt eos}) is produced or maximum length is reached. Receiver's input is an embedding of the message: either the embedding of the single-symbol message, or the last hidden state of the RNN cell that corresponds to the {\tt eos} symbol.

Once Sender and Receiver are defined, the user wraps them into EGG-implemented wrappers which determine the communication and optimization scenarios. Importantly, the actual user-specified Sender and Receiver architectures can be agnostic to whether single-symbol or variable-length communication is used; and to whether Gumbel-Softmax relaxation- or REINFORCE-based training is performed. 
In Figure~\ref{fig:agent_wrappers} we illustrate different communication/training scenarios: (a) single-symbol communication, trained with Gumbel-Softmax relaxation, (b) single-symbol communication, trained with REINFORCE, (c) variable-length communication, trained with Gumbel-Softmax relaxation, (d) variable-length communication, trained with REINFORCE.

Once the Game instance is defined, everything is ready for training. That is, the user has to pass the game instance to {\tt core.Trainer}, as shown in Figure~\ref{fig:trainer}.

We report some results obtained with the code we just described. We used the following parameters. The vision module is a pre-trained LeNet-1~\cite{Lecun1990} instance, the maximal message length is set to 2, the communication between the agents is done through LSTM units with hidden-size 20, vocabulary size is 10. The agents are trained with REINFORCE for 15 epochs with batch size of 32, and the loss is per-pixel cross-entropy. 

\begin{figure}
\centering
\includegraphics[width=0.5\textwidth]{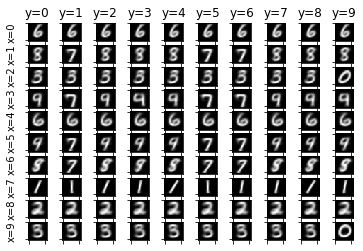}
\caption{The emergent code-book in the MNIST auto-encoder game. After training, we feed all 100 possible two-symbol messages $xy$ from the size-10 vocabulary to Receiver and show the returned images. The rows iterate over the first symbol $x$, the columns enumerate the second symbol, $y$. The {\tt eos} symbol has id 0.}
\label{fig:tutorial}
\end{figure}

In Figure~\ref{fig:tutorial} we illustrate the language that emerges in this setup. To do this, we enumerate all possible 100 two-symbol messages $x,y$ and input them to Receiver. We report all images that Receiver produces. The {\tt eos} symbol is fixed to be 0, hence if the first symbol is 0 then the second symbol is ignored (top row of Figure~\ref{fig:tutorial}).

Note that the first symbol $x$ tends to denote digit identity: $x \in \{2,4, 7,8, 9\}$. In contrast, the second symbol $y$ is either ignored ($x \in \{4, 8\}$) or specifies the style of the produced digit ($x \in \{3, 7\}$). The second symbol has the most striking effect with $x = 7$, where $y$ encodes the rotation angle of the digit 1. 

\section{Some pre-implemented games}
EGG contains implementations of several games. 
They (a) illustrate how EGG can be used to explore interesting research questions, (b) provide reference usage patterns and building blocks, (c) serve as means to ensure reproducibility of studies reported in the literature. For example, EGG incorporates an  implementation of the signaling game of \citet{Lazaridou2016} and \citet{Bouchacourt2018}. It contains code that was recently used to study the communicative efficiency of artificial LSTM-based agents~\cite{Chaabouni2019anti} and the information-minimization properties of emergent discrete codes \cite{Kharitonov2019}.\footnote{A small illustration can be run in Google Colaboratory: \url{https://colab.research.google.com/github/facebookresearch/EGG/blob/master/egg/zoo/language_bottleneck/mnist-style-transfer-via-bottleneck.ipynb}.} Finally, EGG provides a pre-implemented game that allows to train agents entirely via the command line and external input/output files, without having to write a single line of Python code. We hope this will lower the learning curve for those who want to experiment with language emergence without previous coding experience.

\section{Conclusion and future work}
We introduced EGG, a toolkit for research on emergence of language in games. We outlined its main features  design principles. Next, we briefly reviewed how training with a discrete communication channel is performed. Finally, we walked through the main steps for implementing a MNIST autoencoding game using EGG.

We intend to extend EGG in the following directions. First, we want to provide support for multi-direction and multi-step communicative scenarios. Second, we want to add more advanced tooling for analyzing the properties of the emergent languages (such as compositionality; \citealt{Andreas:2019}). We will also continue to enlarge the set of pre-implemented games, to build a library of reference implementations.


\section*{Acknowledgments}
We are grateful to Roberto Dess\`i and Tomek Korbak for their contributions to the EGG codebase and to Serhii Havrylov for sharing his code with us.

\balance
\bibliography{emnlp-ijcnlp-2019,marco}
\bibliographystyle{acl_natbib}

\end{document}